\def\year{2021}\relax
\documentclass[letterpaper]{article} 
\usepackage{aaai21}  
\usepackage{times}  
\usepackage{helvet} 
\usepackage{courier}  
\usepackage[hyphens]{url}  
\usepackage{graphicx} 
\usepackage{natbib}  
\usepackage{caption} 
\usepackage{amsmath}
\usepackage{amssymb}
\usepackage{mathrsfs}
\usepackage{algorithmicx}
\usepackage[linesnumbered]{algorithm2e}
\usepackage{arydshln}

\title{Learning to Continually Learn Rapidly from Few and Noisy Data}
\author{
    Nicholas I-Hsien Kuo$^{1, *}$,
    Mehrtash Harandi$^2$,
    Nicolas Fourrier$^3$,\\
    Christian Walder$^{1, 4}$,
    Gabriela Ferraro$^{1, 4}$,
    Hanna Suominen$^{1, 4, 5}$\\
}

\affiliations{
    $^1:$ RSCS, The Australian National University, Canberra, ACT, Australia\\
    $^2:$ ECSE, Monash University, Melbourne, Victoria, Australia\\
    $^3:$ Research Center of L\'{e}onard de Vinci P\^{o}le Universitaire, Paris La D\'{e}fense, France\\
    $^4:$ Data61, CSIRO, Canberra, ACT, Australia\\
    $^5:$ Department of Computing, University of Turku, Turku, Finland\\
    $^*:$ u6424547@anu.edu.au
}

\begin{document}

\maketitle

\begin{abstract}
Neural networks suffer from catastrophic forgetting and are unable to sequentially learn new tasks without guaranteed stationarity in data distribution. Continual learning could be achieved via replay -- by concurrently training externally stored old data while learning a new task. However, replay becomes less effective when each past task is allocated with less memory. To overcome this difficulty, we supplemented replay mechanics with meta-learning for rapid knowledge acquisition. By employing a meta-learner, which \textit{learns a learning rate per parameter per past task}, we found that base learners produced strong results when less memory was available. Additionally, our approach inherited several meta-learning advantages for continual learning: it demonstrated strong robustness to continually learn under the presence of noises and yielded base learners to higher accuracy in less updates. We released our codes at\\
 \url{https://github.com/Nic5472K/MetaLearnCC2021_AAAI_MetaSGD-CL}.
\end{abstract}

\section{Introduction}
In standard practice, it is assumed that the examples used to train a neural network are drawn independently and identically distributed (i.i.d.) from some fixed distribution. However in a changing environment, \textit{continual learning} is required where an agent faces a continual stream of data, and must adapt to learn new predictions. This remains a difficult challenge because neural networks are known to suffer from  \textit{catastrophic  forgetting}~\cite{mccloskey1989catastrophic}, the phenomenon which they abruptly lose the capability to solve a previously learnt task when information for solving a new task is incorporated.   

It was shown that catastrophic forgetting can be effectively alleviated by concurrently training data of past tasks. These \textit{replay}~\cite{robins1995catastrophic} and \textit{pseudo replay}~\cite{shin2017continual} mechanisms have continuously been improved~\cite{rebuffi2017icarl, lopez2017gradient, chaudhry2018efficient}; and a recent paper from \citet{chaudhry2019tiny} have even shown that forget prevention could be made possible by co-training as little as one single sample of past data per parameteric update. Nonetheless, we found that the replayed data was still required to be sampled from a large set of externally stored memory. Sampling from an insufficient amount of external memory would lead a network to overfit on early tasks thus generalise poorly on subsequent tasks. 

This paper combines the work of \citet{chaudhry2019tiny} with meta-learning for rapid knowledge acquisition. We extended their method with \textit{MetaSGD}~\cite{li2017meta} to maintain high performances under the constraint of learning with low memory consumption. We employed a meta-learner to assist the training procedure of a task-specific base learner of which \textit{learns a learning rate per parameter per past task}. This hybrid approach not only produced strong results when less memory was available, but it also inherited several desirable meta-learning advantages for continual learning. Our approach demonstrated strong robustness against the realistic scenario which we continually learn under the presence of noises, and it could also optimise base learner parameters to achieve higher accuracy in less iterations.

\section{Preliminaries}

Our narrative below mainly follows \citet{lopez2017gradient}, \citet{kirkpatrick2017overcoming}, and \citet{chaudhry2019tiny}.

\subsection{The Framework of Continual Learning}

In continual learning, we employ a task-specific base learner $F$ to sequentially learn the $n$ tasks of $\omega_1, \ldots, \omega_n \in \Omega$ in a 
\textit{continuum} of data
\begin{align}\label{eq:continuum}
\hspace{-0.5mm} \Xi = \{(x_1^1, \omega_1, y_1^1), \ldots, (x_\gamma^i, \omega_i, y_\gamma^i), \ldots, (x_\Gamma^n, \omega_n, y_\Gamma^n)\}
\end{align}
to map inputs $x_\gamma^i$ to labels $y_\gamma^i$ from datasets $(x_\gamma^i, y_\gamma^i)\in \mathcal{D}_{\omega_i}$ for all $\Gamma$ data where $\gamma = 1,\ldots, \Gamma$. The aim is to yield a learner that generalises well on subsequent tasks while it maintains high accuracy for learnt tasks.

\begin{figure}[t]
\centering
\includegraphics[width=0.8\columnwidth]{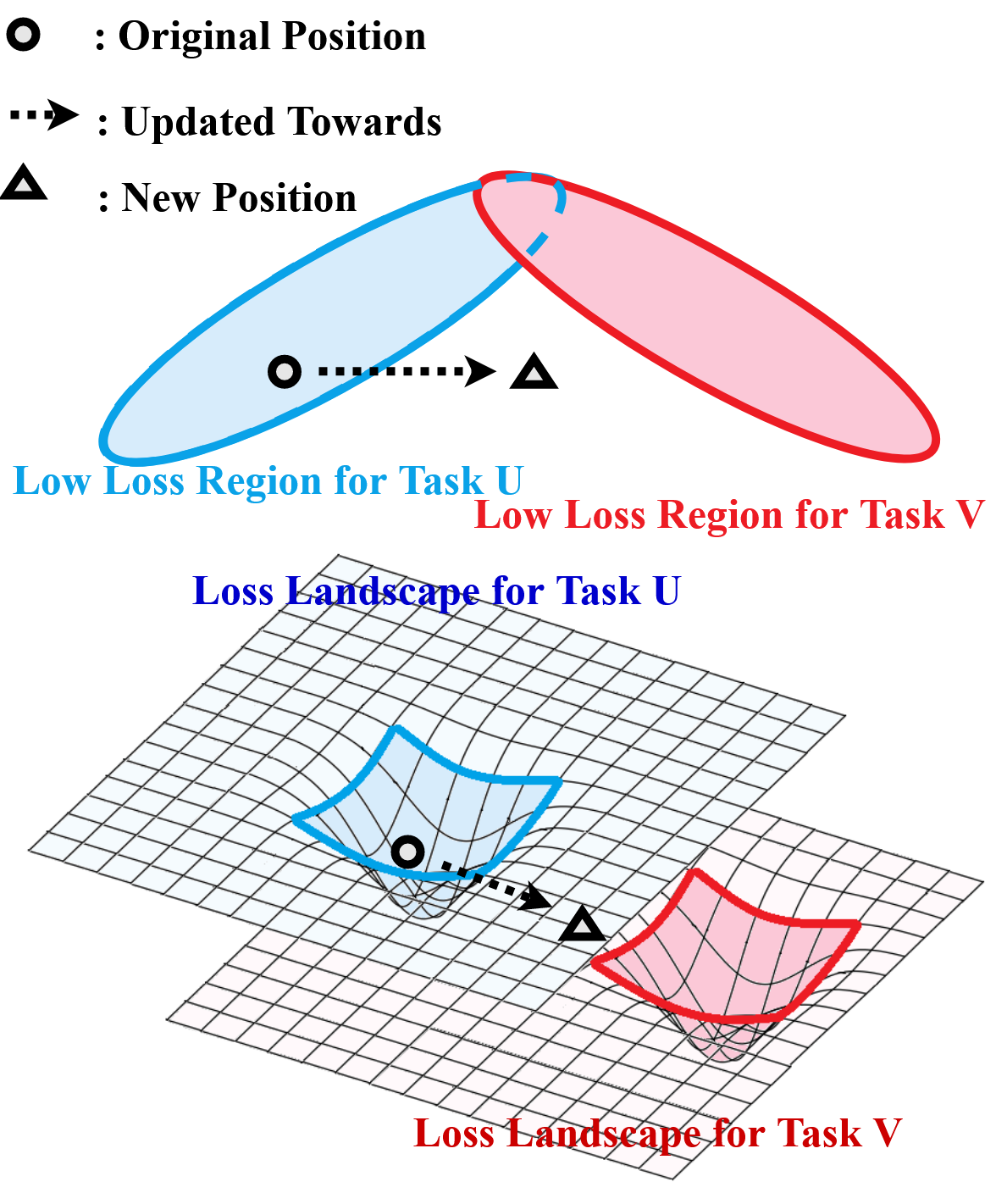}
\caption{Exiting a local minimum while continual learning.}
\label{Fig:01}
\end{figure}

Continual learning is challenging~\cite{kemker2017measuring}. When a learner \textit{only} observes data of a new task, its optimised parameters for an old task will be modified to suit the learning objective of the new task. We elaborate this with the scenario presented in Figure \ref{Fig:01} where a learner optimised for Task U is now sequentially updated for an independent Task V. Na\"ively training the learner with gradient descent will cause the learner to leave the minimum (low loss region) on the loss landscape of Task U in pursuit for that of the new task~\cite{kirkpatrick2017overcoming}\footnote{Inspiration taken from Figure 1 of \citet{kirkpatrick2017overcoming}.}.

\subsection{Replay with Episodic Memory}\label{sec:2.2}

Catastrophic forgetting can be alleviated by replaying all data of past tasks but this inevitably creates a large computational and memory burden. Several studies have shown that memory-based strategy do not require all past training data. \citet{rebuffi2017icarl} prevented forgetting by constructing an exemplar set of past data via learning representation; while \citet{lopez2017gradient} preconditioned the gradients of the current learning task with small \textit{episodic memories} of past tasks. This paper will focus specifically on episodic memory-based methods as \citet{lopez2017gradient}'s original work was continually streamlined by the important works of \citet{chaudhry2018efficient} and \citet{chaudhry2019tiny}.

There are two steps to the episodic memory mechanism. Prior to the sequential training phase, \citet{lopez2017gradient} first define the size of the memory storage $\mathcal{M}_{\omega_i}$ allocated for each task. Then upon training a new task, all stored data from the external memory was used to precondition the gradients of the current task. Though this method yielded very strong results and only stored a section of all past data, it still had difficulties in scaling well as the amount of subsequent tasks increased. 

In their paper, \citet{chaudhry2019tiny} streamlined \citet{lopez2017gradient}'s method with \textit{Experience Replay}~(ER), and showed that catastrophic forgetting can be alleviated without utilising all data stored within episodic memories. Algorithm \ref{alg:ER} shows their ER approach; and the core mechanism of ER is to \textit{sample} mini-batches $B_\mathcal{M}$ from all stored data of past tasks (see line 8) to effectively mitigate the computational burden from employing episodic memories. The sampled content was then concurrently trained with the existing batch of the current learning task and they used SGD to update parameters $\theta$ of base learner $F$ (see line 10). Impressively, they found that strong results for forget prevention can be achieved even when $B_\mathcal{M}$ only sampled 1 data from all memory $\cup\mathcal{M}_{\omega_i}$.

\begin{algorithm}[t]
\fbox{Define memory storage}\\
$\mathcal{M}_{\omega_i}$$ \leftarrow \{\} \hspace{1.5mm} \forall i \text{ such that }\omega_i \in \Omega$\\
\For{$i = 1, \ldots,\lvert \Omega \lvert$}{
    \For{$(x_\gamma^i, \omega_i, y_\gamma^i) \in$ $\Xi$}{
        \fbox{Update memory}\\
        $\mathcal{M}_{\omega_i}\leftarrow\mathcal{M}_{\omega_i}\cup(x_\gamma^i, y_\gamma^i)$\\
        \uIf{$i > 2$}{
            \fbox{Sample from all memory}\\
            $B_\mathcal{M}\sim\cup_{u = 1}^{i-1}\mathcal{M}_{\omega_u}$\\
            \fbox{Update with sampled memory}\\
            $\theta$ $\leftarrow \text{SGD}((x_\gamma^i, y_\gamma^i)\cup B_\mathcal{M}, \theta)$
        }
        \Else{
            \fbox{Conventional update}\\
            $\theta$ $\leftarrow \text{SGD}((x_\gamma^1, y_\gamma^1), \theta)$
        }
    }
}
\caption{Experience replay recipe}\label{alg:ER}
\end{algorithm}

\subsection{Learning with Episodic Memory Sampled\\
from a Tiny Memory Storage}\label{sec:2.3}
To employ episodic memory, the practitioner would need to explicitly define the size of the external memory storage; \\
\textit{but what are the disadvantages of selecting small sizes?}\\
A small memory storage meant that the replay mini-batch $B_\mathcal{M}$ would have a higher chance of re-sampling the same past data for concurrent training (see line 11 of Algorithm \ref{alg:ER}). A base learner would hence overfit on earlier tasks through the consecutive re-exposure of a small set of past data and thus underfit on future tasks. This issue was lightly investigated in \citet{chaudhry2019tiny}\footnote{
See Section 5.5 and Appendix C of \citet{chaudhry2019tiny}.}, but they only compared medium-size to large-size memories\footnote{
See Table 7 in \citet{chaudhry2019tiny}, where they compared 1000 to 20000 memory units. What happens when we use 100?\label{FN:03}}.
In this current work, we limit the memory size to only a handful of data per past task to study its consequences for continual learning and propose meta-learning as a remedy.

\section{Meta-learning}
Meta-learning studies rapid knowledge acquisition and has contributed to few shot learning~\cite{ravi2016optimization} where a base learner is required to learn from a low resource environment. In this paper, we formulate continual learning with tiny episodic memory storage as a low resource problem. Though sequential learning does not lack data on subsequent tasks, our base learner is still required to learn from a small memory unit for past tasks.

\subsection{Learning to Optimise with MetaSGD}

One archetype of meta-learning focuses on optimisation. This type of approach is also known as \textit{learning to learn}~\cite{andrychowicz2016learning}. Optimisation is cast as the learning problem of a meta-learner of which is used to configure the weights of the base learner network. It replaces the classic gradient descent of
\begin{align}
    \theta_{t+1} &=\theta_t - \alpha \nabla_\theta \mathcal{L}_{t+1}(\theta_t) \label{eq:GD}
\end{align}
for a base learner $F$ with parameters $\theta_t$ at time $t$; where $\alpha$ is the learning rate and that $\nabla_\theta \mathcal{L}_{t+1}(\theta_t)$ is the gradient of the learner loss with respect to the parameters.

One particularly simple method called \textit{MetaSGD}~\cite{li2017meta} proposes to update learner parameters via
\begin{align}
\theta_{t+1} &= \theta_t - \beta \odot \nabla_\theta \mathcal{L}_{t+1}(\theta_t), \label{eq:MetaSGD}
\end{align}
where $\beta\in\mathbb{R}^{\lvert\theta_t\lvert}$ and that symbol $\odot$ represents the element-wise product. Learning rates $\beta$ are the meta-learning target of MetaSGD, and hence a unique learning rate is learnt for every parameter to enable important features to be updated more quickly. In their paper, \citet{li2017meta} showed that MetaSGD achieved strong results for few shot learning; and \citet{nicholas2020m2sgd} showed that MetaSGD could converged base learners much more rapidly than SGD\footnote{
See Figure 5 of \citet{nicholas2020m2sgd} where MetaSGD-like methods decreased updates from 37500 to 10000 steps.}. 
We will now simplify our notation by denoting $\nabla_\theta \mathcal{L}_{t+1}(\theta_t)$ as $\nabla \mathcal{L}$.

\subsection{An Analysis on the Update of Experience Replay for Continual Learning}

Under conventional setup, loss $\mathcal{L}$ aggregates every individual cost $C$ therein a mini-batch $B_n$ with
\begin{align}
\mathcal{L} &= \frac{1}{\lvert B_n \lvert} \sum_{i = 1}^{\lvert B_n \lvert}C(\hat{y_i}, y_i)
\end{align}
where $\hat{y_i}$ is the prediction of learner $F$ and that $y_i$ is the corresponding true label. Hence under the ER regime, every batch used in continual learning will be substituted with $B_n \leftarrow B^*$ where $B^*=B_n\cup B_{\mathcal{M}}$ (see line 11 of Algorithm \ref{alg:ER}). Then due to the linearity of summation, the new loss $\mathcal{L}^*$ can be written as
\begin{align}\label{eq:pg3_2}
\mathcal{L}^* &= \frac{1}{\lvert B^* \lvert} \left[ \sum_{i = 1}^{\lvert B_n \lvert}C(\hat{y_i}, y_i) + \sum_{j = 1}^{\lvert B_\mathcal{M} \lvert}C(\hat{y_j}, y_j) \right].
\end{align}

If a base learner had observed $V$ tasks, we can highlight task specificity and rewrite Equation \eqref{eq:pg3_2} as
\begin{align}\label{eq:pg4_1}
    \mathcal{L}^* &= 
        \frac{1}{\lvert B^* \lvert} \left[ \sum_{i = 1}^{\lvert B_n \lvert}C(\hat{y_i}, y_i) + 
        \sum_{u = 1}^{V - 1}\sum_{j = 1}^{\lvert B_{\mathcal{M}_u} \lvert}C(\hat{y_j}, y_j) \right]
\end{align}
where $B_\mathcal{M} = \cup_{u = 1}^{V - 1} B_{\mathcal{M}_u}$. Then with $\mathbb{L}$ as the task-wise loss we have 
\begin{align}\label{eq:pg4_2}
    \mathcal{L}^* &= 
        \frac{1}{\lvert B^* \lvert} 
        \left[
        \lvert B_n \lvert \mathbb{L}_V +
        \sum_{u = 1}^{V - 1} \lvert B_{\mathcal{M}_u} \lvert \mathbb{L}_u
        \right];
\end{align}
and this will create the standard parametric update of
\begin{multline}\label{eq:pg4_no3}
    \theta_{t+1} =
        \theta_t - 
        \left(
            \frac{\alpha}{\lvert B^* \lvert}
            \lvert B_n \lvert
        \right) 
            \nabla \mathbb{L}_V \\
            - 
        \sum_{u = 1}^{V - 1}
        \left(
            \frac{\alpha}{\lvert B^* \lvert}
            \lvert B_{\mathcal{M}_u} \lvert
        \right)
            \nabla \mathbb{L}_u
\end{multline}
of which we could employ MetaSGD and introduce $\beta$ as
\begin{align}\label{eq:main}
    \theta_{t+1} &=
        \theta_t - 
            \beta_V \odot
            \nabla \mathbb{L}_V - 
        \sum_{u = 1}^{V - 1}
            \beta_u \odot
            \nabla \mathbb{L}_u
\end{align}
to \textit{learn a learning rate per parameter per past task}.

\begin{figure}[t]
\centering
\includegraphics[width=0.6\columnwidth]{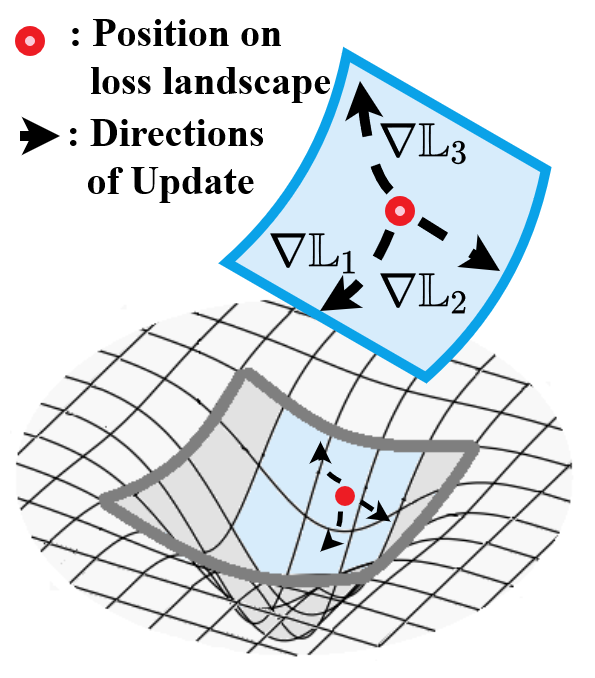}
\caption{Task-wise directional updates.}
\label{Fig:02}
\end{figure}

To elaborate, consider the scenario in Figure \ref{Fig:02} where a learner observes its third task. The learner faces different directions of update $\nabla\mathbb{L}$ from the competing learning objectives of different tasks. \textit{Interference}~\cite{riemer2018learning} hence occurs and transfer in knowledge is difficult; this is further complicated as learner $F$ could easily surpass thousands of parameters in practice \cite{szegedy2015going}. Thus instead of applying handcrafted rules such as SGD to update a base learner, we employ \textbf{MetaSGD for Continual Learning} (MetaSGD-CL) to explore a stream of changes in a series of task structures.

\section{Implementation Details on MetaSGD-CL}
Upon learning a new task $V$ (including task 1), MetaSGD-CL initialises a trainable vector $b_V = 0.01\cdot\tilde{\mathbf{1}}\in\mathbb{R}^{\lvert \theta \lvert}$. In the context of Equation \eqref{eq:main}, we set
\begin{align}\label{eq:ind}
    \beta_{V_k} &=
        \text{ReLU}(
            \text{min}(
                b_{V_k}, \kappa
            )
        );
\end{align}
where $\kappa$ is a postive constant hyper-parameter. Thus every individual learning rate $\beta_{V_k}$ lies within the range of $[0, \kappa]$. In this paper, we default $\kappa$ as 0.02 unless specified otherwise.

When sequential learning a new task $V$ (excluding task 1), MetaSGD-CL freezes past learning rates $\beta_{u}$ for all $u < V$ and updates base learner $F$ with the formulation below
\begin{align}\label{eq:practical}
    \theta_{t+1} &=
        \theta_t - 
            \beta_V \odot
            \nabla \mathbb{L}_V - 
        \frac{1}{V-1}
        \sum_{u = 1}^{V - 1}
            \beta_u \odot
            \nabla \mathbb{L}_u.
\end{align}

\begin{figure*}[t]
\centering
\includegraphics[width=0.8\textwidth]{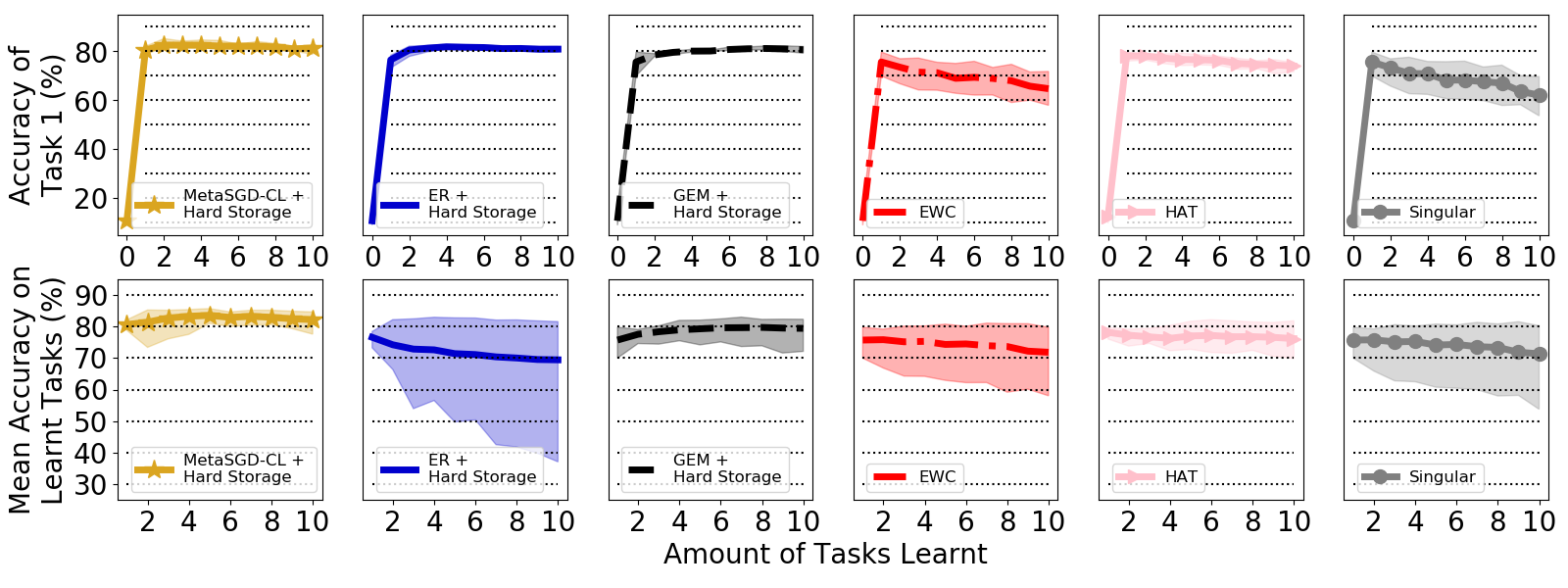} 
\caption{Performances for permuted MNIST. Top row to reflect catastrophic forgetting; and bottom row to reflect overfitting.}
\label{Fig:03}
\end{figure*}

Equation (\ref{eq:practical}) accounts for alignments in $\nabla\mathbb{L}$ due to similarity among arbitrary task structures; $\frac{1}{V-1}$ hence prevents the over-parametrisation in any given feature dimension. 

The loss function of MetaSGD-CL is adapted from \citet{andrychowicz2016learning}
\begin{multline}\label{eq:MetaSGDLoss}
    \mathscr{L}(\beta_V) = 
        \mathcal{L}
            ( 
                F_{\theta_{t+1}}(x_t), y_t
                \lvert \beta_V,\\ 
                        \{
                        \beta_u \text{ for }
                            u = 1, \ldots, (V - 1)
                        \}
            ),
\end{multline}
based on an updated learner $F_{\theta_{t+1}}$ with original data $x_t$. Thus, the objective for MetaSGD-CL is to further fine-tune learner $F$. The meta-learning targets $\beta$s are updated with Adam~\cite{kingma2014adam} with learning rate 0.01.

\section{Experiments}

This section includes some benchmarks, demonstrates the benefits of MetaSGD-CL over ER, and an ablation study.

\subsection{Dataset}

We conducted experiments on \textit{permuted MNIST}~\cite{kirkpatrick2017overcoming} and \textit{incremental Cifar100}~\cite{lopez2017gradient}. Previously, \citet{lopez2017gradient} and \citet{chaudhry2019tiny} showed that memory-based approaches outperformed continual learning techniques of other archetypes and achieved state of the art status. However, we will show that when the amount of memory decreases, ER performances can drop sharply on the considerably simple permuted MNIST tasks.

MNIST~\cite{lecun1998gradient} contains 60,000 training and 10,000 test images in 10 classes of 28$\times$28 grey-scale handwritten digits. Permuted MNIST first reshaped each image as a vector of 784(=28$\times$28) pixels, then applied a unique permutation on all vectors of that task. In our experiments, we sequentially trained over 10 permuted MNIST tasks.

Cifar100~\cite{krizhevsky2009learning} is a coloured dataset with 100 classes of objects. There are 50,000 training and 10,000 test images all presented on 32$\times$32 pixel boxes. In our incremental class experiments, we sequentially introduced 5 classes per task drawn without replacement from the dataset. We reserved the first 50 classes for pre-training a classifier, and tested on the remaining 50 classes, hence there were 10 tasks in total. See the next subsection for more details.

\begin{table}[t]
\centering
\begin{tabular}{|l||l|l|}
    \hline
    \textbf{Methods} & \textbf{FA1} (\%) & \textbf{ACC} (\%) \\
    \hline
    \hline
    MetaSGD-CL (Ours) & \textbf{81.02} & \textbf{82.19} \\
    \hline
    ER & 80.60 & 69.42 \\
    \hline
    GEM & 80.71 & 79.43 \\
    \hline
    EWC & 64.80 & 71.84 \\
    \hline
    HAT & 74.03 & 76.17 \\
    \hline
    Singular & 62.18 & 71.34 \\
    \hline
\end{tabular}
\caption{Metrics for permuted MNIST in Figure \ref{Fig:03}.}
\label{Tab:01}
\end{table}

\subsection{Base Learners}

For permuted MNIST, we followed \citet{lopez2017gradient} and trained MLPs as base learner. As for incremental Cifar100, we followed \citet{mai2020batch}\footnote{\citet{mai2020batch} won the CVPR CLVision 2020 challenge.} and used their best practice which employed a fixed feature extractor and only sequentially trained the classifier networks. We pre-trained a ResNet18~\cite{he2016deep} as feature extractor on the first 50 classes of Cifar100; then updated MLP base learners as classifiers. All MLP base learners of our study had 2 hidden layers with 100 dimensions followed by a softmax layer.

Following \citet{lopez2017gradient}, we used batch size 10. Each task of permuted MNIST observed 100 batches; and it was 1 epoch each for incremental Cifar100. Our MetaSGD-CL was described in the last section; and all baselines were updated with SGD with learning rate 0.01.

\subsection{Metrics}

The 2 metrics of which we use in this paper are based on a matrix $\mathscr{R}\in\mathbb{R}^{n\times n}$ of which we employ to store all accuracy during the entire course of continual learning. Row $i$ recorded the accuracy after training on the $i$th dataset; and column $m$ recorded the accuracy of the $m$th task.

The severity of catastrophic forgetting can be reflected via\\
\underline{\textbf{Final Accuracy of Task 1}} \hspace{1.75mm}(FA1) \hspace{4.5mm}$\mathscr{R}_{n,1}$, \hspace{12.5mm}; while\\
\underline{\textbf{Average accuracy}} \hspace{12.5mm}(ACC) \hspace{3mm}$\frac{1}{n}\sum_{i = 1}^{n}\mathscr{R}_{n, i} $\\
can show the overall performance and the severity of overfitting of the base learner. For both of our metrics, it is the larger the better.

\begin{figure}[t]
\centering
\includegraphics[width=0.9\columnwidth]{ZZ_Azz05_Overall.png}
\caption{Performances for incremental Cifar100.$^{\text{6}}$}
\label{Fig:04}
\end{figure}

\begin{table}[t]
\centering
\begin{tabular}{|l||l|l|}
    \hline
    \textbf{Methods} & \textbf{FA1} (\%) & \textbf{ACC} (\%) \\
    \hline
    \hline
    MetaSGD-CL (Ours) & \textbf{81.52} & \textbf{81.06} \\
    \hline
    ER & 79.28 & 79.50 \\
    \hline
    GEM & 79.92 & 80.88 \\
    \hline
    EWC & 67.46 & 71.98 \\
    \hline
    HAT & 71.98 & 78.86 \\
    \hline
    Singular & 62.86 & 73.27\\
    \hline
\end{tabular}
\caption{Metrics for incremental Cifar100 in Figure \ref{Fig:04}.}
\label{Tab:02}
\end{table}

\subsection{Continual Learning Techniques}

Our benchmark experiments included 6 setups.
\textit{Singular} was the scenario where we na\"ively sequentially trained a base learner. Then, we tested our \textit{MetaSGD-CL} against the baseline \textit{ER}~\cite{chaudhry2019tiny}. In addition, we considered the memory-based \textit{Gradient Episodic Memory}~(GEM) of \citet{lopez2017gradient}, the regularisation-based \textit{Elastic Weight Consolidation}~(EWC) of \citet{kirkpatrick2017overcoming}, and the dynamic architectural-based \textit{Hard Attention to the Task}~(HAT) of \citet{serra2018overcoming}. 

\subsection{Remark 1: Preventing Catastrophic Forgetting}
Our results on permuted MNIST are in Figure \ref{Fig:03} and Table \ref{Tab:01}; and those for incremental Cifar100 are in Figure 4\footnote{
Refer to our codes (link in Abstract) to reformat Figure \ref{Fig:04} as Figure \ref{Fig:03} for further inspection.\label{FN4}}
and Table 2. Base on the FA1 metric in the 2 tables, we validated \citet{lopez2017gradient} and \citet{chaudhry2019tiny}'s finding such that memory-based approaches were more effective in preventing catastrophic forgetting.

Our memory-based approach used \textit{Hard Storage}~\cite{lopez2017gradient}, which assigned 250 memory units to each tasks. For each update, the less computationally efficient GEM used all stored data to prevent forgetting; while both MetaSGD-CL and ER randomly sampled 10 data from each task-wise storage to append them as $B_\mathcal{M}$ to the original batch $B_n$ (see Equation \eqref{eq:pg3_2}). Samples of old tasks do not change throughout training.

The base learners of HAT had hidden dimensions of 400 instead. With only 100 hidden dimensions, HAT could not prevent forgetting and behaved as that in \citet{ahn2019uncertainty}. 

\subsection{Remark 2: 3 Advantages of Meta-SGD over ER}

\begin{figure}[h]
\centering
\includegraphics[width=1\columnwidth]{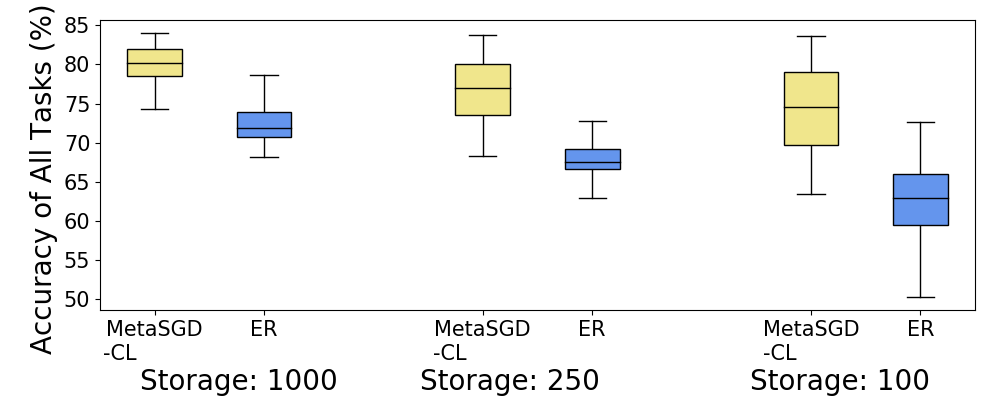}
\caption{Ring buffer of different sizes for permuted MNIST.}
\label{Fig:05}
\end{figure}

\subsubsection{Remark 2.1: Does not overfit on old tasks}

From Figure \ref{Fig:03} and Table \ref{Tab:01}, ER was comparable to MetaSGD-CL in FA1 but much lower in ACC. This served as a clear indication of ER's overfitting problem. For incremental Cifar100 and Table \ref{Tab:02}, \citet{mai2020batch}'s experimental setup limited some extent of overfitting by only continually training the classifier network. However, the bottom subplot of Figure \ref{Fig:04} showed that ER's variability (in blue) was much larger than MetaSGD-CL's (in yellow); and hence overfitting persisted.

We further tested overfitting with the alternative memory storage of \textit{Ring Buffer}~\cite{chaudhry2019tiny}. Ring buffer shares $\Phi$ memory units \textbf{for all tasks}\footnote{If $\Phi = 250$ and that $n = 10$ tasks, then it is $25$ units per task.}. By design, the buffer is under-utilised in the early phases of sequential learning; and that stored samples do not change. Compared to hard storage, ring buffer is much more computationally efficient. However, it is also more likely to incur overfitting because data of all past tasks are now sharing the same buffer. 

Figure \ref{Fig:05} shows the results for MetaSGD-CL and ER on permuted MNIST with ring buffers of 1000, 250, and 100 units. Note, these settings were much smaller than the scenarios tested in \citet{chaudhry2019tiny}$^{\text{3}}$. Similar to Remark 1, each iteration randomly sampled 10 past data as $B_\mathcal{M}$ but from one unified storage (see line 11 of Algorithm~\ref{alg:ER}).

Unsurprisingly, the performances of either methods decreased as the sizes of memory dropped. This was a natural consequence of overfitting by repeatedly re-sampling from a small set of past data. ER is one of the state of the art techniques; but still, there was a significant amount of lesion on the considerably simple permuted MNIST benchmark when the size of ring buffer changed from 1000 to 100. MetaSGD-CL was more capable of maintaining higher accuracy.

\subsubsection{Remark 2.2: Rapid Knowledge Acquisition}
We also tested 2 realistic continual learning scenarios. First, we considered when there was a limited amount of source data; and second, when data was collected from a noisy environment.

\begin{figure}[t]
\centering
\includegraphics[width=\columnwidth]{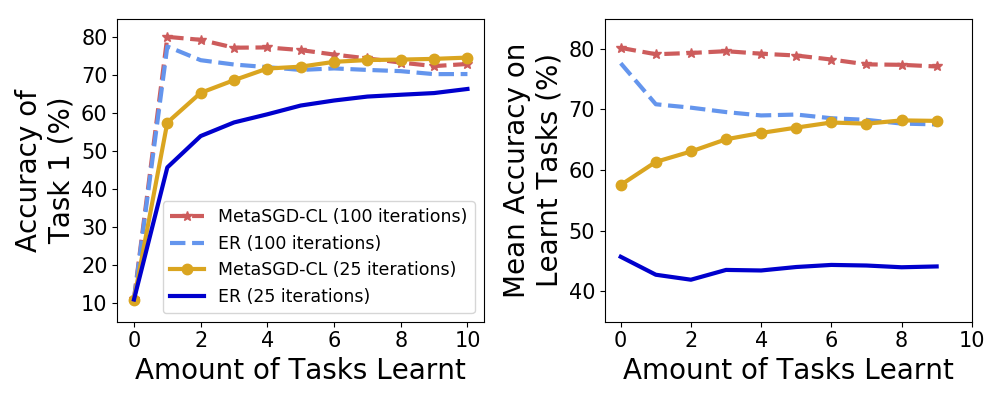}
\caption{Permuted MNIST but with iteration reduction.}
\label{Fig:06}
\end{figure}

In Figure \ref{Fig:06}, we tested MetaSGD-CL and ER on permuted MNIST with ring buffer but reduced the available training data from 100 iterations to 25 iterations per task. Unsurprisingly, the test accuracy had dropped; in addition, overfitting continued to occur in ER. Our MetaSGD-CL approach was able to leverage on meta-learning and it yielded higher accuracy in few iterations; also, its mean accuracy concurrently increased with its accuracy for task 1.

\begin{figure}[t]
\centering
\includegraphics[width=\columnwidth]{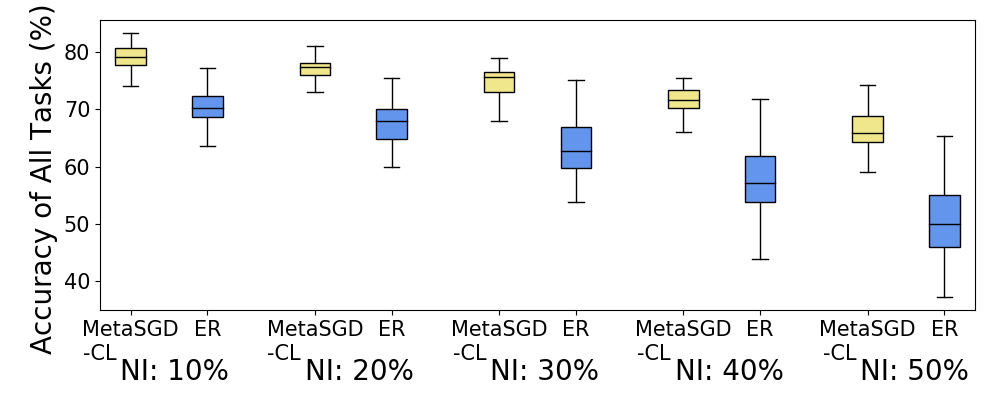}
\caption{Permuted MNIST but with noise injection.}
\label{Fig:07}
\end{figure}

\subsubsection{Remark 2.3: Strong Robustness to Noise} In Figure \ref{Fig:07}, we tested MetaSGD-CL and ER on permuted MNIST with ring buffer, and we demonstrated the consequences of \textit{noise injection} (NI) in both the training data and stored memory. NI randomly shuffled a section of pixels in the MNIST images. Our lowest setting injected 10\% noise, and the highest setting injected up to 50\% noise.

We found that MLPs trained with MetaSGD-CL were less susceptible to NI than their counterparts trained with ER; there were higher accuracy and smaller variability. This was likely because the learning rates $\beta$ of Equation~(\ref{eq:main}) assigned higher learning rates to features that were less prune to NI.

\subsection{Remark 3: Behaviours on Learning Rate $\beta$}
\subsubsection{Remark 3.1: Convergence in $\beta$ values} As we previously mentioned, the maximal value of $\beta$ was defaulted as $\kappa = 0.02$ (see Equation \eqref{eq:ind}). The value for $\kappa$ was purposely set low because we hypothesised that there existed conflicted objectives between meta-learning and continual learning. 

As shown in Figure \ref{Fig:06}, meta-learning is naturally capable of rapid knowledge acquisition. That is, we hypothesised that $\beta$ would grow unboundedly with large $\kappa$ values; and this would hence indirectly cause overfitting on old tasks. 

\begin{figure}[t]
\centering
\includegraphics[width=\columnwidth]{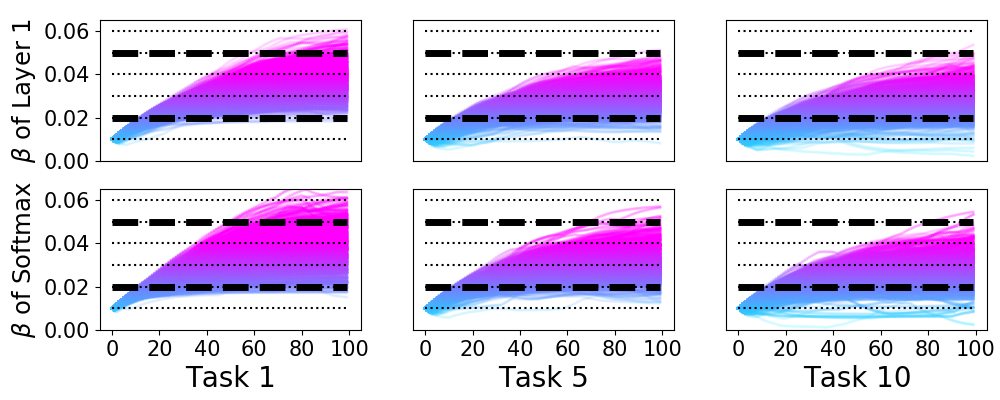}
\caption{Transition in task-wise $\beta$ magnitudes.}
\label{Fig:08}
\end{figure}

To our surprise, $\kappa = 0.1$ yielded 76.95 ACC and this was a minor lesion to $\kappa = 0.02$'s 82.19 ACC shown in Table \ref{Tab:01}. We found that it was because that extreme $\beta$ values converged as the base learner observed more tasks. As shown in Table \ref{Tab:03} and Figure \ref{Fig:08}, only a minor proportion of $\beta$ values were lower than $0.02$ and larger than $0.05$. We found that as the base learner observed more tasks, the proportion of large magnitude task-wise $\beta$ decreased, while that of small magnitude task-wise $\beta$ increased. That is, less parametric updates were made to the base learner from data of later tasks. 

Furthermore, a larger proportion of large values were found in the task-wise $\beta$ of deeper base learner layers. Thus MetaSGD-CL prioritised in updating the weights of the deeper layers rather than those of the shallower layers.

\begin{table}[t]
\begin{tabular}{|l||r|r|r|r|r|}
\hline
\multicolumn{6}{|l|}{Proportion of $\beta > 0.05$ (in \%) in task-wise learning rate}\\
\hline
\textbf{Tasks} &
    \textbf{2} & 
    \textbf{4} & 
    \textbf{6} & 
    \textbf{8} &
    \textbf{10}\\
\hline
\hline
Layer 1 &
    1.39 & 
    0.63 & 
    0.47 & 
    0.46 &
    0.36\\
\hline
Layer 2 &
    2.26 & 
    1.41 & 
    1.33 & 
    1.32 &
    1.30\\
\hline
Softmax Layer &
    3.64 & 
    2.28 & 
    1.88 & 
    1.76 &
    1.74\\
\hline
\hline
\multicolumn{6}{|l|}{Proportion of $\beta < 0.02$ (in \%) in task-wise learning rate}\\
\hline
\textbf{Tasks} &
    \textbf{2} & 
    \textbf{4} & 
    \textbf{6} & 
    \textbf{8} &
    \textbf{10}\\
\hline
\hline
Layer 1 &
    1.43 & 
    3.90 & 
    6.56 & 
    8.31 &
    8.79\\
\hline
Layer 2 &
    4.29 & 
    6.59 & 
    7.71 & 
    10.51 &
    11.27\\
\hline
Softmax Layer &
    6.02 & 
    7.50 & 
    8.38 & 
    8.92 &
    11.58\\
\hline
\end{tabular}
\caption{Ratios of extreme task-wise $\beta$s with $\kappa = 0.1$.}
\label{Tab:03}
\end{table}

\begin{table}[ht]
\centering
\begin{tabular}{|l||l|l|}
    \hline
    \textbf{Methods} & \textbf{FA1} (\%) & \textbf{ACC} (\%) \\
    \hline
    \hline
    Un-ablated MetaSGD-CL & 81.02 & 82.19 \\
    \hdashline
    Singular & 62.18 & 71.34 \\
    \hdashline
    Replace all old $\beta$ with 0s & 69.81 & 74.41 \\
    \hline
    Replace all old $\beta$ with 0.01s & 73.48 & 77.16 \\
    \hline
    Replace all old $\beta$ with 0.1s & 72.57 & 75.80 \\
    \hline
\end{tabular}
\caption{Metrics for permuted MNIST in Figure \ref{Fig:03}.}
\label{Tab:04}
\end{table}

\subsubsection{Remark 3.2: Ablation Study}

To further understand MetaSGD-CL, we reported 3 ablation studies in Table \ref{Tab:04}. When learning a new task, we removed the learnt learning rates $\beta$, and replaced them with 0, 0.01, and 0.1.

When $\beta$ of the old tasks were replaced with 0, catastrophic forgetting occurred. However, these ablated performances were still much better than the na\"ively implemented sequential learning. Interestingly, when $\beta$ of the old tasks were replaced with non-zero values, the ablated performances were similar. We attributed this to formulation Equation \eqref{eq:main}; as long as information of the past were given, MetaSGD-CL was able to learn appropriate $\beta$ for the new task to indirectly balance base learner parametric configurations for all tasks.

\section{Related Studies and Future Work}
Like EWC and \textit{Learning without Forgetting}~(LwF)~\cite{li2017learning}, MetaSGD-CL also alleviated catastrophic forgetting via modifications in the backward pass. Thus, it can likely be modified to prevent forgetting via regularisation.

Alternative meta-learning approaches have been employed in continual learning. \citet{riemer2018learning} combined GEM with Reptile~\cite{nichol2018first}. While \citet{javed2019meta} continually learned via representation learning with \textit{Model-Agnostic Meta-Learning}~(MAML)~\cite{finn2017model}-like techniques. A large scale comparison for such hybrid approaches would thus be an interesting future study.

\section{Conclusion}
This paper introduced \textbf{MetaSGD-CL} as an extension to \citet{chaudhry2019tiny}'s \textit{ER} technique to remedy the overfitting on old tasks when memory-based continual learning were allocated with small memory sizes. MetaSGD-CL leveraged on meta-learning by \textit{learning a learning rate per parameter per past task}. It alleviated catastrophic forgetting and also prevented base learners from overfitting and achieved high performances for all tasks. In addition, MetaSGD-CL optimised base learners in fewer iterations, and showed higher robustness against noises in training data.

\section{Acknowledgments}
This research was supported by the Australian Government Research Training Program (AGRTP) Scholarship. We also thank our reviewers for the constructive feedback.

\bibliography{main}

\begin{thebibliography}{26}
\providecommand{\natexlab}[1]{#1}
\providecommand{\url}[1]{\texttt{#1}}
\providecommand{\urlprefix}{URL }
\expandafter\ifx\csname urlstyle\endcsname\relax
  \providecommand{\doi}[1]{doi:\discretionary{}{}{}#1}\else
  \providecommand{\doi}{doi:\discretionary{}{}{}\begingroup
  \urlstyle{rm}\Url}\fi

\bibitem[{Ahn et~al.(2019)Ahn, Cha, Lee, and Moon}]{ahn2019uncertainty}
Ahn, H.; Cha, S.; Lee, D.; and Moon, T. 2019.
\newblock Uncertainty-Based Continual Learning with Adaptive Regularisation.
\newblock In \emph{Neural Information Processing Systems}.

\bibitem[{Andrychowicz et~al.(2016)Andrychowicz, Denil, Gomez, Hoffman, Pfau,
  Schaul, Shillingford, and De~Freitas}]{andrychowicz2016learning}
Andrychowicz, M.; Denil, M.; Gomez, S.; Hoffman, M.~W.; Pfau, D.; Schaul, T.;
  Shillingford, B.; and De~Freitas, N. 2016.
\newblock Learning to Learn by Gradient Descent by Gradient Descent.
\newblock In \emph{Neural Information Processing Systems}.

\bibitem[{Chaudhry et~al.(2018)Chaudhry, Ranzato, Rohrbach, and
  Elhoseiny}]{chaudhry2018efficient}
Chaudhry, A.; Ranzato, M.; Rohrbach, M.; and Elhoseiny, M. 2018.
\newblock Efficient Lifelong Learning with A-GEM.
\newblock In \emph{International Conference on Learning Representations}.

\bibitem[{Chaudhry et~al.(2019)Chaudhry, Rohrbach, Elhoseiny, Ajanthan,
  Dokania, Torr, and Ranzato}]{chaudhry2019tiny}
Chaudhry, A.; Rohrbach, M.; Elhoseiny, M.; Ajanthan, T.; Dokania, P.~K.; Torr,
  P.~H.; and Ranzato, M. 2019.
\newblock On Tiny Episodic Memories in Continual Learning.
\newblock In \emph{International Conference on Machine Learning Workshop}.

\bibitem[{Finn, Abbeel, and Levine(2017)}]{finn2017model}
Finn, C.; Abbeel, P.; and Levine, S. 2017.
\newblock Model-Agnostic Meta-Learning for Fast Adaptation of Deep Networks.
\newblock In \emph{International Conference on Machine Learning}.

\bibitem[{He et~al.(2016)He, Zhang, Ren, and Sun}]{he2016deep}
He, K.; Zhang, X.; Ren, S.; and Sun, J. 2016.
\newblock Deep Residual Learning for Image Recognition.
\newblock In \emph{Computer Vision and Pattern Recognition}.

\bibitem[{Javed and White(2019)}]{javed2019meta}
Javed, K.; and White, M. 2019.
\newblock Meta-Learning Representations for Continual Learning.
\newblock In \emph{Neural Information Processing Systems}.

\bibitem[{Kemker et~al.(2018)Kemker, McClure, Abitino, Hayes, and
  Kanan}]{kemker2017measuring}
Kemker, R.; McClure, M.; Abitino, A.; Hayes, T.; and Kanan, C. 2018.
\newblock Measuring Catastrophic Forgetting in Neural Networks.
\newblock In \emph{AAAI Conference on Artificial Intelligence}.

\bibitem[{Kingma and Ba(2015)}]{kingma2014adam}
Kingma, D.~P.; and Ba, J. 2015.
\newblock Adam: A Method for Stochastic Optimisation.
\newblock In \emph{International Conference on Learning Representations}.

\bibitem[{Kirkpatrick et~al.(2017)Kirkpatrick, Pascanu, Rabinowitz, Veness,
  Desjardins, Rusu, Milan, Quan, Ramalho, Grabska-Barwinska
  et~al.}]{kirkpatrick2017overcoming}
Kirkpatrick, J.; Pascanu, R.; Rabinowitz, N.; Veness, J.; Desjardins, G.; Rusu,
  A.~A.; Milan, K.; Quan, J.; Ramalho, T.; Grabska-Barwinska, A.; et~al. 2017.
\newblock Overcoming Catastrophic Forgetting in Neural Networks.
\newblock In \emph{Proceedings of the National Academy of Sciences}. National
  Acad Sciences.

\bibitem[{Krizhevsky(2009)}]{krizhevsky2009learning}
Krizhevsky, A. 2009.
\newblock Learning Multiple Layers of Features from Tiny Images.
\newblock In \emph{University of Toronto Tech Report}.

\bibitem[{Kuo et~al.(2020)Kuo, Harandi, Fourrier, Walder, Ferraro, and
  Suominen}]{nicholas2020m2sgd}
Kuo, N.; Harandi, M.; Fourrier, N.; Walder, C.; Ferraro, G.; and Suominen, H.
  2020.
\newblock M2SGD: Learning to Learn Important Weights.
\newblock In \emph{Computer Vision and Pattern Recognition Workshop on
  Continual Learning in Vision}.

\bibitem[{LeCun et~al.(1998)LeCun, Bottou, Bengio, and
  Haffner}]{lecun1998gradient}
LeCun, Y.; Bottou, L.; Bengio, Y.; and Haffner, P. 1998.
\newblock Gradient-Based Learning Applied to Document Recognition.
\newblock In \emph{Proceedings of the IEEE}.

\bibitem[{Li and Hoiem(2017)}]{li2017learning}
Li, Z.; and Hoiem, D. 2017.
\newblock Learning without Forgetting.
\newblock In \emph{Pattern Analysis and Machine Intelligence}.

\bibitem[{Li et~al.(2017)Li, Zhou, Chen, and Li}]{li2017meta}
Li, Z.; Zhou, F.; Chen, F.; and Li, H. 2017.
\newblock Meta-SGD: Learning to Learn Quickly for Few-Shot Learning.
\newblock \emph{arXiv preprint arXiv:1707.09835} .

\bibitem[{Lopez-Paz and Ranzato(2017)}]{lopez2017gradient}
Lopez-Paz, D.; and Ranzato, M. 2017.
\newblock Gradient Episodic Memory for Continual Learning.
\newblock In \emph{Neural Information Processing Systems}.

\bibitem[{Mai et~al.(2020)Mai, Kim, Jeong, and Sanner}]{mai2020batch}
Mai, Z.; Kim, H.; Jeong, J.; and Sanner, S. 2020.
\newblock Batch-level Experience Replay with Review for Continual Learning.
\newblock In \emph{Computer Vision and Pattern Recognition Workshop}.

\bibitem[{McCloskey and Cohen(1989)}]{mccloskey1989catastrophic}
McCloskey, M.; and Cohen, N.~J. 1989.
\newblock Catastrophic Interference in Connectionist Networks: The Sequential
  Learning Problem.
\newblock In \emph{Psychology of Learning and Motivation}.

\bibitem[{Nichol, Achiam, and Schulman(2018)}]{nichol2018first}
Nichol, A.; Achiam, J.; and Schulman, J. 2018.
\newblock On First-Order Meta-Learning Algorithms.
\newblock \emph{arXiv preprint} .

\bibitem[{Ravi and Larochelle(2017)}]{ravi2016optimization}
Ravi, S.; and Larochelle, H. 2017.
\newblock Optimisation as a Model for Few-Shot Learning.
\newblock In \emph{International Conference on Learning Representations}.

\bibitem[{Rebuffi et~al.(2017)Rebuffi, Kolesnikov, Sperl, and
  Lampert}]{rebuffi2017icarl}
Rebuffi, S.-A.; Kolesnikov, A.; Sperl, G.; and Lampert, C.~H. 2017.
\newblock iCaRL: Incremental Classifier and Representation Learning.
\newblock In \emph{Computer Vision and Pattern Recognition}.

\bibitem[{Riemer et~al.(2018)Riemer, Cases, Ajemian, Liu, Rish, Tu, and
  Tesauro}]{riemer2018learning}
Riemer, M.; Cases, I.; Ajemian, R.; Liu, M.; Rish, I.; Tu, Y.; and Tesauro, G.
  2018.
\newblock Learning to Learn Without Forgetting by Maximizing Transfer and
  Minimizing Interference.
\newblock In \emph{International Conference on Learning Representations}.

\bibitem[{Robins(1995)}]{robins1995catastrophic}
Robins, A. 1995.
\newblock Catastrophic Forgetting, Rehearsal and Pseudorehearsal.
\newblock In \emph{Connection Science}.

\bibitem[{Serra et~al.(2018)Serra, Suris, Miron, and
  Karatzoglou}]{serra2018overcoming}
Serra, J.; Suris, D.; Miron, M.; and Karatzoglou, A. 2018.
\newblock Overcoming Catastrophic Forgetting with Hard Attention to the Task.
\newblock In \emph{International Conference on Machine Learning}.

\bibitem[{Shin et~al.(2017)Shin, Lee, Kim, and Kim}]{shin2017continual}
Shin, H.; Lee, J.~K.; Kim, J.; and Kim, J. 2017.
\newblock Continual Learning with Deep Generative Replay.
\newblock In \emph{Neural Information Processing Systems}.

\bibitem[{Szegedy et~al.(2015)Szegedy, Liu, Jia, Sermanet, Reed, Anguelov,
  Erhan, Vanhoucke, and Rabinovich}]{szegedy2015going}
Szegedy, C.; Liu, W.; Jia, Y.; Sermanet, P.; Reed, S.; Anguelov, D.; Erhan, D.;
  Vanhoucke, V.; and Rabinovich, A. 2015.
\newblock Going Deeper with Convolutions.
\newblock In \emph{Computer Vision and Pattern Recognition}.

\end{thebibliography}

\end{document}